\documentclass{article}

\usepackage{arxiv}

\usepackage[utf8]{inputenc} 
\usepackage[T1]{fontenc}    
\usepackage{hyperref}       
\usepackage{url}            
\usepackage{booktabs}       
\usepackage{amsfonts}       
\usepackage{nicefrac}       
\usepackage{microtype}      
\usepackage{lipsum}		
\usepackage{graphicx}
\usepackage{natbib}
\usepackage{doi}

\title{The Evolution of Darija Open Dataset: Introducing Version 2}

\author{ 
{\hspace{1mm}Aissam Outchakoucht}\\
	\texttt{aissam.outchakoucht@gmail.com} \\
	\And
{\hspace{1mm}Hamza Es-Samaali} \\
	\texttt{hamza.essamaali@gmail.com} \\
}

\renewcommand{\headeright}{}
\renewcommand{\undertitle}{}
\renewcommand{\shorttitle}{}

\hypersetup{
pdftitle={A template for the arxiv style},
pdfauthor={Aissam Outchakoucht, Hamza Es-Samaali},
pdfkeywords={Dataset, Darija, dialect, open-source, NLP, Low resource languages},
}

\begin{document}
\maketitle

\begin{abstract}
	Darija Open Dataset (DODa) represents an open-source project aimed at enhancing Natural Language Processing capabilities for the Moroccan dialect, Darija. With approximately 100,000 entries, DODa stands as the largest collaborative project of its kind for Darija-English translation. The dataset features semantic and syntactic categorizations, variations in spelling, verb conjugations across multiple tenses, as well as tens of thousands of translated sentences. The dataset includes entries written in both Latin and Arabic alphabets, reflecting the linguistic variations and preferences found in different sources and applications.
    The availability of such dataset is critical for developing applications that can accurately understand and generate Darija, thus supporting the linguistic needs of the Moroccan community and potentially extending to similar dialects in neighboring regions.
    This paper explores the strategic importance of DODa, its current achievements, and the envisioned future enhancements that will continue to promote its use and expansion in the global NLP landscape.
\end{abstract}


\section{Introduction}
The Darija Open Dataset (DODa), now in its second version, represents a significant advancement in the field of Natural Language Processing (NLP) for the Moroccan dialect, Darija. This dataset has been expanded from 10,000 entries \citep{doda} to now include more than 100,000 translated entries, featuring a broader array of sentences and increased lexical content, and now incorporates both Arabic and Latin scripts to reflect the dual-script usage prevalent in contemporary Moroccan communication.\newline
The primary motivation for developing DODa was the observed scarcity of linguistic resources available for Darija, a dialect spoken by millions \citep{laoudi} yet markedly underrepresented in digital and linguistic research. The decision to develop a second version of the DODa and to document its evolution in this data paper was driven by two significant advancements in the dataset's development. First, the sheer volume of data in DODa has expanded more than tenfold compared to its initial version. Second, the inclusion of Arabic script entries alongside the original Latin alphabet entries addresses the linguistic realities of Darija speakers and researchers. This dual-script approach not only enhances the dataset’s scale but also makes it more accessible and useful for a wider range of applications and studies. These enhancements underscore our commitment to continuously improve DODa, making it an invaluable resource for the NLP community.\newline
Translation in NLP is crucial. It serves as a vital bridge that extends low resource languages to the broad array of resources and sophisticated applications developed for globally dominant languages like English. In fact, while training a large language model directly on raw Darija text has its merits, such as promoting linguistic authenticity and specific cultural nuances, translation offers a more expansive benefit. It allows Darija to leverage existing NLP technologies and infrastructures, bypassing many developmental hurdles. This approach accelerates the availability of high-quality NLP tools for Darija speakers in addition to ensuring that Darija is well-integrated into the global digital landscape. That is why, in DODa, our focus extends beyond merely collecting raw Darija text, this opens up the dataset to a wider academic and technological audience, increasing the scope of research and the potential for innovative applications.

\section{Related works}

In computational linguistics for Moroccan Darija, few datasets exist that are conducive to advanced research and application development. This section reviews significant efforts in this domain. The Multi Arabic Dialect Applications and Resources (MADAR) project \citep{madar} is notable for its wide coverage of 25 Arabic city dialects including Fez and Rabat from Morocco. MADAR provides a parallel corpus specific to the travel domain, encompassing 14,000 entries, alongside a lexicon containing 3,063 words. Notably, Darija expressions in MADAR utilize the Arabic alphabet and are translated into Modern Standard Arabic (MSA).\newline
Another resource is the Moroccan Dialect Electronic Dictionary (MDED) \citep{mded}, which features over 12,000 entries, also transcribed using Arabic characters and translated into MSA. This dictionary is based predominantly on MSA content and includes a substantial corpus of approximately 34,000 sentences sourced from various origins, tailored for Darija-to-MSA translation.\newline
Additional resources include a Moroccan Darija Wordnet \citep{wordnet}, constructed using a bilingual Moroccan-English dictionary, and various online repositories that provide translated terminology from specific sectors such as gastronomy. Nonetheless, these resources often face limitations in scope, dynamics, and accessibility. Many are not open-source, lack continuous updates or community involvement, and/or are narrowly focused on specific domains. The scientific rigor and documentation of these datasets are frequently inadequate, rendering them less viable for scholarly citation and use in rigorous research projects. This gap underscores the need for datasets that are not only comprehensive and dynamic but also systematically documented and readily accessible to the research community.\newline
Another noteworthy project, the Offensive Moroccan Comments Dataset (OMCD) \citep{omcd}, focuses on detecting offensive language within the Moroccan dialect, comprising 8,024 comments. Additionally, the Goud dataset \citep{goud} offers a rich collection of 158,000 articles and corresponding headlines from the Goud.ma news site, written in Arabic script. These headlines are exclusively in Moroccan Darija, whereas the articles may be in pure Darija, MSA, or a combination thereof.\newline
Finally, a new initiative in the field of Darija linguistics is AtlasIA \citep{atlasia}, an open-source project that extends beyond data collection to include the development of translation models and other NLP tools for Darija. It already has 3 open sourced models in HuggingFace. Currently, AtlasIA encompasses around 45,000 sentences, significantly enriching the computational linguistics landscape for Moroccan Darija. This project collaborates closely with DODa; a considerable portion of AtlasIA’s data has been sourced from DODa. AtlasIA supports DODa's efforts by presenting untranslated sentences from DODa on their web application, enabling contributions from non-technical volunteers. This synergy leverages the strengths of both datasets and fosters a broader community involvement in the enhancement of NLP resources for Moroccan Darija.

\section{Description of the dataset}

DODa is not just a lexicon, nor a collection of translated sentences; it is a comprehensive linguistic resource designed to enhance research and application development in the Moroccan dialect. This dataset transcends the conventional boundaries of language databases by providing a rich compilation of entries that include:

\begin{itemize}
    \item Semantic and Syntactic Categorization: Words in the dataset are classified not only by their semantic category such as food, economy, education, etc; but also by syntactic function, distinguishing between verbs, nouns, adjectives, etc.
    \item Linguistic Variations: DODa captures the diversity of the Darija dialect through multiple spelling variants for the same word, verb-to-noun transformations, and masculine-to-feminine gender adaptations.
    \item Verb Conjugations: The dataset includes an extensive set of verb conjugations across various tenses.
    \item Sentence Collection: Beyond individual words, DODa boasts a vast corpus of over 45,000 translated sentences, providing valuable context and usage examples for the language.
    \item Ongoing Contributions: DODa is dynamically expanding, with over 43,000 Darija sentences currently under translation by our contributors. This process of continual updates helps the dataset grow and evolve, reflecting the dedicated involvement of our community of 23 contributors.
\end{itemize}

\begin{table}
	\caption{DODa Translation progress update}
	\centering
	\begin{tabular}{lll}
		\toprule
		Translated words     & Translated sentences     & Ongoing \\
		\midrule
		56,894 & 45,378  & $\sim$43,000 \\     
		\bottomrule
	\end{tabular}
	\label{tab:table}
\end{table}

Additionally, the dataset includes specialized files that cater to unique linguistic needs:

\begin{itemize}
    \item Idioms and Expressions: Separate files are dedicated to idioms and uniquely Moroccan expressions, enriching the dataset's cultural relevance.
    \item Name Databases: Files containing lists of Moroccan male and female names enhance the personalization and localization of NLP applications.
    \item Lexical Correspondences: Special files document correspondences between different lexical forms, such as verbs and their nominal forms (e.g., read/reading), supporting advanced linguistic analysis.
\end{itemize}

\begin{figure}
    \centering
    \includegraphics[width=8cm]{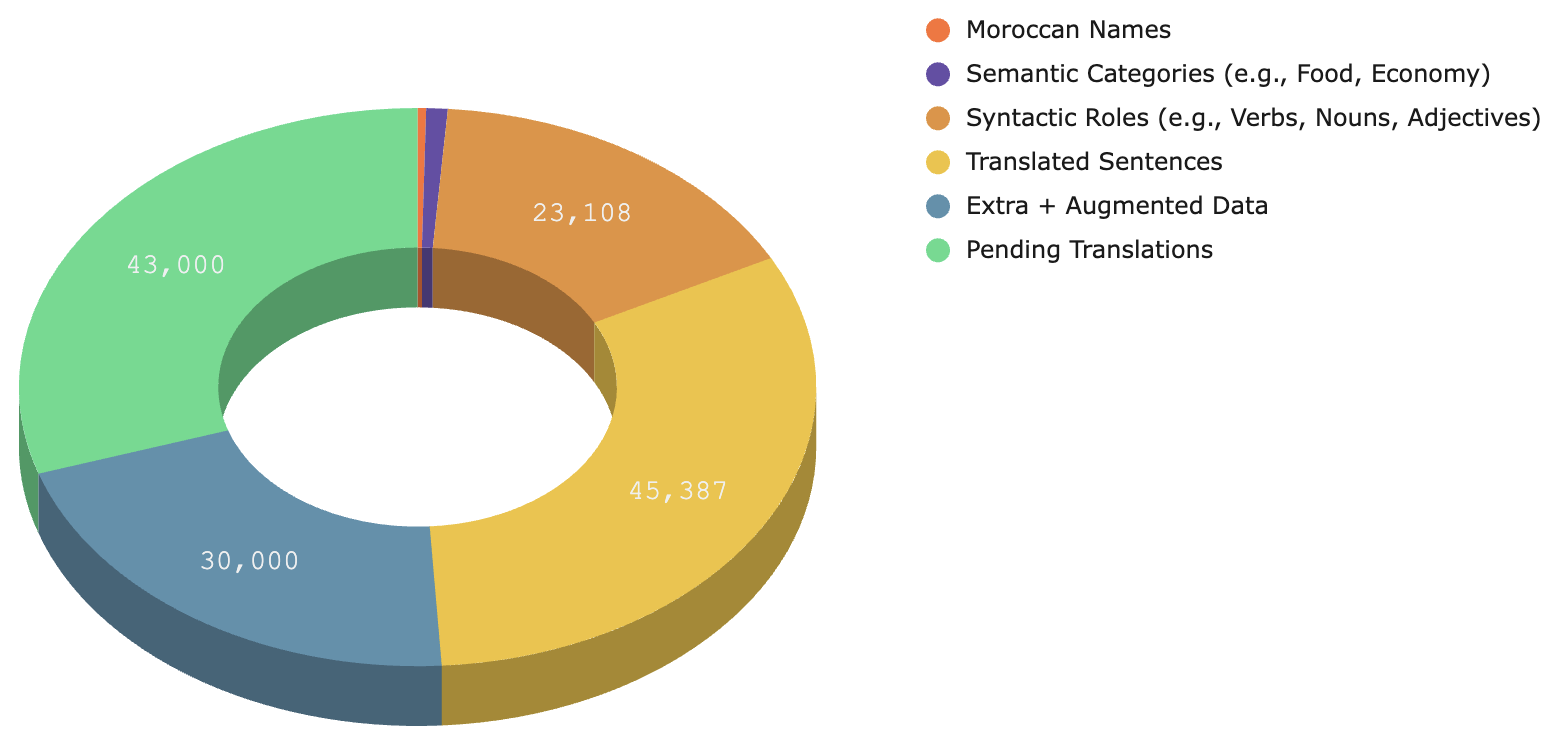}
    \caption{Distribution of Content Types in the DODa}
    \label{fig:galaxy}
\end{figure}

Table 1 and Figure 1 illustrate the diverse range of content types within DODa. They highlight the dataset's extensive focus on syntactic roles, with 23,108 entries, and translated sentences, which account for 45,387 entries. These two categories form the core of DODa, facilitating in-depth linguistic analysis and translation applications. The chart also shows significant ongoing efforts in translation, with 43,000 sentences awaiting processing, underscoring the dataset's dynamic and evolving nature. Additionally, it includes 30,000 entries under extra and augmented data. Smaller but vital components include 1,245 entries categorized by semantic themes like art, education and food, and 493 entries listing Moroccan names, which help personalize and localize NLP applications. This distribution showcases the broad and multifaceted approach of DODa in addressing the linguistic needs of Moroccan Darija.

\subsection{Technical Specifications and User Guide}

The dataset is meticulously structured to ensure ease of access and integration into NLP applications. It is hosted on GitHub, where users can freely download the data in various formats, contribute to the ongoing translation efforts, or utilize the provided APIs for seamless integration into their projects. A detailed user guide on the platform offers instructions on how to contribute to or use the dataset, ensuring that both new and experienced users can navigate and make the most of DODa effectively.

\subsection{Vision for Research and Application}

The Darija Open Dataset is not just a collection of data; it is a gateway to pioneering new research pathways in the field of NLP. By providing a dataset with such depth and breadth, DODa encourages researchers to explore novel aspects of the Moroccan dialect, such as comparing linguistic structures across scripts (Arabic vs. Latin) and analyzing the impact of dialectal variations on language processing technologies. The ongoing expansion of the dataset promises to continually enhance its utility and relevance in the global NLP landscape.

\subsection{License}

DODa is open sourced for everyone under the Creative Commons Attribution-NonCommercial 4.0 International License. This licensing framework is chosen to align with academic and research purposes, giving a slight advantage to open source applications of the dataset.\newline
This default non-commercial stipulation allows for greater control and ethical use of DODa. By restricting commercial usage without explicit permission, we try to ensure that the dataset adheres to ethical standards and is used in ways that align with the foundational goals of promoting Moroccan linguistic heritage and technology. It fosters a community-driven development model, where researchers, linguists, and language enthusiasts can contribute and collaborate, enhancing the dataset’s richness and utility without the overriding influence of commercial interests.\newline
While DODa does not preclude the use of its data in commercial applications, it does require that such use be negotiated separately. This approach is not about financial gain but ensures that commercial entities contribute back to the community, potentially through collaborations, enhancements of the dataset, or other forms of support that enrich the DODa project. This strategy aims to maintain a balance between wide accessibility for educational and research purposes and responsible commercial use that also supports the dataset's growth and sustainability.

\subsection{Future directions}

In addition to the ongoing efforts to expand and diversify the data within DODa, we are also developing an evaluation dataset specifically designed to assess the performance of translation models for Moroccan Darija. This evaluation dataset will provide standardized metrics and benchmarks that are crucial for objectively measuring the accuracy and reliability of various NLP tools, particularly those focused on translation.\newline
The evaluation dataset is intended to include a wide range of test cases covering different aspects of the Darija language, from basic vocabulary to complex sentence structures, ensuring comprehensive testing of translation models. It will be designed to challenge the models' ability to handle idiomatic expressions, contextual nuances, and the syntactic diversity of Moroccan Darija. By providing a robust framework for testing, this evaluation dataset will help developers identify strengths and weaknesses in their models, guiding further improvements.

\section{Conclusion}

DODa has emerged as a cornerstone in the advancement of NLP for the Moroccan dialect, Darija. This dataset, now enriched with around 100,000 entries spanning various linguistic facets, not only serves as the most extensive resource of its kind but also underscores the crucial role of open-source projects in bridging linguistic gaps. The development of DODa, particularly in its second iteration, has set a new benchmark in the field by expanding its scope and depth, thus facilitating more sophisticated and tailored NLP applications. This expansion is also qualitative, enhancing the dataset's capability to support advanced research and the development of technologies that can perform complex tasks such as sentiment analysis, real-time translation, and AI-driven contextual responses.\newline
We acknowledge and thank the contributors who have played a crucial role in the development and expansion of DODa. Their efforts have been instrumental in reaching this new milestone. As DODa continues to grow, it is poised to facilitate more effective NLP applications and to serve as a model for similar initiatives for other underrepresented languages.

\bibliographystyle{unsrtnat}
\bibliography{references}  

\begin{thebibliography}{8}
\providecommand{\natexlab}[1]{#1}
\providecommand{\url}[1]{\texttt{#1}}
\expandafter\ifx\csname urlstyle\endcsname\relax
  \providecommand{\doi}[1]{doi: #1}\else
  \providecommand{\doi}{doi: \begingroup \urlstyle{rm}\Url}\fi

\bibitem[Outchakoucht and Es-Samaali(2021)]{doda}
Aissam Outchakoucht and Hamza Es-Samaali.
\newblock Moroccan dialect -darija- open dataset.
\newblock In \emph{arXiv preprint arXiv:2103.09687}, 2021.

\bibitem[J.~Laoudi and Voss(2018)]{laoudi}
L.~Donatelli S.~Tratz J.~Laoudi, C.~Bonial and C.~Voss.
\newblock Towards a computational lexicon for moroccan darija: Words, idioms, and constructions.
\newblock In \emph{Proceedings of the Joint Workshop on Linguistic Annotation, Multiword Expressions and Constructions (LAW-MWE-CxG-2018)}, pages 74--85, 2018.

\bibitem[H.~Bouamor(2018)]{madar}
et~al. H.~Bouamor.
\newblock The madar arabic dialect corpus and lexicon.
\newblock In \emph{11th International Conference on Language Resources and Evaluation. LREC, European Language Resources Association (ELRA),}, pages 3387--3396, 2018.

\bibitem[R.~Tachicart(2014)]{mded}
H.~Jaafar R.~Tachicart, K.~Bouzoubaa.
\newblock Building a moroccan dialect electronic dictionary (mded).
\newblock In \emph{5th International Conference on Arabic Language Processing CITALA, Oujda, Morocco}, 2014.

\bibitem[K.~Mrini(2017)]{wordnet}
F.~Bond K.~Mrini.
\newblock Building the moroccan darija wordnet (mdw) using bilingual resources.
\newblock In \emph{Proceedings of the International Conference on Natural Language, Signal and Speech Processing (ICNLSSP)}, 2017.

\bibitem[Essefar~K.(2023)]{omcd}
El~Mahdaouy A. et~al. Essefar~K., Ait Baha~H.
\newblock Omcd: Offensive moroccan comments dataset.
\newblock In \emph{Lang Resources and Evaluation 57}, page 1745–1765. Springer, 2023.
\newblock \doi{10.1007/s10579-023-09663-2}.

\bibitem[A.~Issam(2022)]{goud}
K.~Mrini A.~Issam.
\newblock Goud.ma: A news article dataset for summarization in moroccan darija.
\newblock In \emph{AfricaNLP workshop at ICLR}, 2022.

\bibitem[Community(2024)]{atlasia}
Moroccan Community.
\newblock Atlasia.
\newblock In \emph{\url{https://huggingface.co/datasets/atlasia/darija-translation}}, 2024.

\end{thebibliography}






\end{document}